%% file: neurips_2021.tex
\documentclass{article}

     \PassOptionsToPackage{numbers, compress}{natbib}

\usepackage[preprint]{neurips_2021}

\usepackage[utf8]{inputenc} 
\usepackage[T1]{fontenc}    
\usepackage{hyperref}       
\usepackage{url}            
\usepackage{booktabs}       
\usepackage{amsfonts}       
\usepackage{nicefrac}       
\usepackage{microtype}      
\usepackage{xcolor}         
\usepackage{diagbox}
\usepackage{Definitions}
\usepackage{multirow}
\usepackage{wrapfig}

\title{Fairness for Robust Learning to Rank}

\author{%
  Omid Memarrast\\
  Department of Computer Science\\
  University of Illinois at Chicago\\
  Chicago, IL 60607 \\
  \texttt{omemar2@uic.edu} \\
  \And
  Ashkan Rezaei\\
  Department of Computer Science\\
  University of Illinois at Chicago\\
  Chicago, IL 60607 \\
  \texttt{arezae4@uic.edu} \\
  \And
  Rizal Fathony\\
  Bosch Center for Artificial Intelligence\\
  Carnegie Mellon University\\
  Pittsburgh, PA 15222 \\
  \texttt{rizal@fathony.id} \\
  \And 
  Brian Ziebart\\
  Department of Computer Science\\
  University of Illinois at Chicago\\
  Chicago, IL 60607 \\
  \texttt{bziebart@uic.edu}
}

\begin{document}

\maketitle

\begin{abstract}
  While conventional ranking systems focus solely on maximizing the utility of the ranked items to users, fairness-aware ranking systems additionally try to balance the exposure for different protected attributes such as gender or race. To achieve this type of group fairness for ranking, we derive a new ranking system based on the first principles of distributional robustness. 
We formulate a minimax game between a player choosing a distribution over rankings to maximize utility while satisfying fairness constraints against an adversary seeking to minimize utility while matching statistics of the training data. We show that our approach provides better utility for highly fair rankings than existing baseline methods.
\end{abstract}

\section{Introduction}
\label{intro}
Searching for relevant information through large amounts of data is a ubiquitous
computing task.
Applications include ordering search results (e.g., Google, Bing, or Baidu), personalizing social networks  (e.g., Facebook, Instagram or Twitter), product recommendations for e-commerce sites (e.g., Amazon or eBay), and content recommendation for news/media sites (e.g., YouTube or Netflix). Ranking a subset of items is a crucial component in these applications to help users find relevant information quickly amongst vast amounts of data. 

Rankings often have social implications beyond the immediate utility they provide, since higher rankings provide opportunities for individuals and groups associated with the ranked items. As a consequence, biases in ranking systems---whether intentional or not---raise ethical concerns 
about their long-term economic and societal harming effect \cite{noble2018algorithms}.
Rankings that solely maximize utility or relevance can perpetuate existing societal biases that exist in training data whilst remaining oblivious to the societal detriment they cause by amplifying such biases \cite{o2016weapons}.

Conventional ranking algorithms typically produce rankings to best serve the interests of those conducting searches by ordering the items by the probability of relevance so that utility to the users will be maximized \cite{robertson1977probability}. Users are fulfilled, yet being oblivious to certain attributes of items to be ranked can have a harmful effect on minority groups in the items. Consequently, this could lead to further disparities, particularly for socially salient sub-populations, due to historic and current discriminatory practices 
which has introduced biases into data-driven models \cite{friedman1996bias}. Biased outcomes drawn by these models negatively impact items in marginalized protected groups in critical decision making systems such as hiring or housing where items compete for exposure and being unfair towards one group can lead to winner-takes-all dynamics that reinforce existing disparities \cite{singh2018fairness}. 
Protected group definitions vary between different applications, and can include characteristics such as race, gender, religion, etc. In group fairness, algorithms divide the population into groups based on the protected attribute and guarantee the same treatment for members across groups. In ranking, this treatment can be evaluated using  statistical metrics defined for measuring fairness. In this paper, we mainly focus on two prevalent group fairness measures, \emph{demographic parity} and \emph{disparate treatment} \cite{singh2018fairness}. 

Fair ranking approaches with group fairness properties can be categorized into \emph{post-processing} and \emph{in-processing} methods. \emph{Post-processing} techniques are used to re-rank a given high utility ranking to incorporate fairness constraints while seeking to retain high utility. \citet{singh2018fairness} introduces a re-ranking algorithm that can be optimized to allocate exposure for protected groups based on their group size (i.e., demographic parity) or their merit (i.e., disparate treatment and disparate impact). Following probabilistic ranking principles (PRP), this algorithm optimizes over doubly-stochastic matrices with fairness constraints. Since the solution is a doubly-stochastic matrix, Birkhoff-von Neumann  \cite{birkhoff1940lattice} decomposition is used to infer a stochastic ranking policy from the solution. In a related post-processing method, \citet{biega2018equity} uses integer programming to dynamically optimize amortized equity of attention over time by reordering consecutive rankings. 

These \emph{post-processing} methods assume true relevance labels are available and require other fairness unaware learning methods (e.g., regression) to predict the true labels as a pre-processing step. Recovering from unfair regression based rankings in the re-ranking step may not be feasible in some circumstances \cite{yadav2019fair}.

The fair ranking problem can also be addressed as an \emph{in-processing},  learning-to-rank (LTR) task where the algorithm learns to maximize utility subject to fairness constraints from training data. LTRs often produce probability rankings, which lead to fairness guarantees not on a single ranking problem but on average. DELTR \cite{zehlike2020reducing} extends the ListNet's \cite{cao2007learning} algorithm and proposes an LTR method that  optimizes a weighted summation of a loss function and a fairness criterion. The loss function is a Cross Entropy loss designed for ranking and the fairness objective is a squared hinge loss based on disparate exposure. This algorithm is constrained in how it measures fairness: it only considers the top-1 place in each ranking but not how additional items are ranked. To address this issue, Fair-PG-LTR \cite{singh2019policy}, another fair LTR method, prioritizes utility and fairness simultaneously for the full ranking by making use of a policy gradient optimization algorithm. 

While providing a fairness-utility trade-off, fair LTR approaches need to be robust to outliers and noisy data. For example, the label of recidivism in the COMPAS dataset is regarded to be noisy \cite{eckhouse2017big}. This makes prediction while incorporating fairness constraints more difficult. We aim to solve this problem by constructing a fair robust LTR approach. 


In this paper, we derive a new ranking system based on the first principles of distributional robustness. 
We formulate a minimax game with the \emph{ranker player} choosing a distribution over rankings constrained to provide fairness while maximizing utility and an \emph{adversary player} choosing a distribution of item relevancies that minimizes utility while being similar to training data properties. We show that our approach is able to trade-off between utility and fairness much better at high levels of fairness than existing baseline methods.

\section{Related Works}
\paragraph{Fairness}

In recent years, diverse approaches have been introduced to achieve fair classification according to principles of group fairness. Such approaches can be narrowly classified as input data pre-processing alterations or fair representation learning \cite{Kamiran2012, Calmon2017,Zemel13,feldman2015certifying,gordaliza2019obtaining,donini2018empirical, guidotti2018survey}, in-processing approaches that learn to minimize the prediction loss while incorporating the fairness constraints into the training process \cite{donini2018empirical,zafar2017aistats,zafar2017fairness,zafar2017parity,cotter2019optimization,goel2018non,woodworth2017learning,kamishima2011fairness,bechavod2017penalizing, rezaei2020fairness}, reduction-based methods \cite{agarwal2018reductions,cotter2019optimization}, 
generative-adversarial training \cite{madras2018learning, zhang2018mitigating, celis2019improved,xu2018fairgan,adel2019one} or meta-algorithms \cite{celis2019classification,menon2018cost}.
\paragraph{Fairness in Ranking}
Prior work on fair ranking has been centered on the definitions that rely on the entire list of items for a certain query \cite{zehlike2017fa,celis2018ranking,biega2018equity, singh2018fairness, zehlike2020reducing, gorantla2020ranking}.
In terms of their fairness criteria, these approaches can be considered \textit{unsupervised} where they need average exposure for top ranking items to be equal for all protected groups \cite{singh2018fairness, celis2018ranking, zehlike2020reducing} or \textit{supervised} where they need to predict relevance of items and make average exposure proportional to average utility across groups \cite{biega2018equity, singh2018fairness, sapiezynski2019quantifying, morik2020controlling}. 
Some previous work has focused on composition-based fairness for items maintaining statistical parity where the objects are positioned \cite{yang2017measuring, zehlike2017fa, celis2018ranking,stoyanovich2018online,asudeh2019designing, geyik2019fairness, celis2020interventions}. 
Other works on item fairness base their fairness constraints on statistical parity for pairwise ranking across item groups \cite{beutel2019fairness,kallus2019fairness,narasimhan2020pairwise}.
In addition to item-based approaches, two-sided fair ranking techniques satisfy fairness constraints for both users and items \cite{patro2020fairrec,basu2020framework, patro2020incremental}.

\section{Learning Fair Robust Ranking}
\label{approach}

\subsection{Probabilistic Ranking}

To formulate the ranking problem, we consider a dataset of ranking problems $\Dcal =\{\Rcal^i\}_{i=1}^N$ for $N$ different queries, 
where each $\Rcal^i = \{d_j\}_{j=1}^M$ is a candidate item set of size $M$ for a single query. 
For every item $d_j$ in this set, we denote $\rel(d_j)$ as its corresponding relevance judgment. To formulate the general ranking problem under fairness constraints, we denote the utility of a ranking (permutation) $\pi$ for a single query as $\textrm{Util}(\pi)$. 
The optimization problem can be written as:
$\pi^*  =  \argmax_{\pi \in  \Pi_{\text{fair}}} \textrm{Util}(\pi)$. Utility measures used for rankings are based on the relevance of the individual items being ranked. 
In information retrieval tasks, the utility of 
single ranking problem 
$\Rcal_{|\text{query}} = \{d_j\}_{j=1}^M$ can be expressed in the following general form:
\begin{align}
    \textrm{Util}(\pi) = \sum_{j=1}^M \util_j \expsr_{\pi_j},
    \label{eq:4}
\end{align}
where $\util_j$ represents the utility of a single item $d_j$ based on its relevance $\rel(.)$ and $\expsr$ demonstrates the degree of attention that item $d_j$ gets by being placed at rank $k$ by permutation $\pi$, i.e., $\pi_{d_j}=k$, and $\Gamma_\text{fair}$ is the set of fair probabilistic rankings. 
Note that $\util$ and $\expsr$ are two functions that can be chosen based on the ranking application. Choosing $\util_j=2^{\rel(d_j)}-1$ and $\expsr_k = \frac{1}{\log(1+k)}$ yields the Discounted Cumulative Gain (DCG) \cite{jarvelin2002cumulated}, which is  common evaluation measure in ranking systems. It can be written in the form of \eqref{eq:4} as:
\begin{align}
    \dcg(\pi) = \sum_{d_j \in \Rcal} \frac{2^{\rel(d_j)}-1}{\log(1+\pi_j)}.
\end{align}
The space of all permutations of items is exponential in the number of items, making na\"ive methods that find a utility-maximizing ranking subject to fairness constraints intractable. 
To overcome this problem, we consider a probabilistic ranking in which instead of a single ranking, 
a distribution over rankings 
is used. We define the probability of positioning item $d_j$ at rank $k$ as $P_{j,k}$. Then $\Prob$ constructs a doubly stochastic matrix of size $M\times M$ where entries in each row and each column must sum up to 1. 
By employing the idea of probabilistic ranking, we express the ranking utility in \eqref{eq:4} as an expected utility of a probabilistic ranking:
\begin{align}
 \textrm{U}(\Prob) = \sum_{j = 1}^M\sum_{k=1}^M \Prob_{j,k} \: \util_j \: \expsr_k.
 \label{eq:6}
\end{align}

If we assume the number of items and rank positions are both $M$, we can rewrite \eqref{eq:6} in a vectorized format where $\uvec$ and $\vvec$ are both column vector of size $M$. In this setting the expected utility for probabilistic ranking $\Prob$ is:
    $\textrm{U}(\Prob) = \vec{u}^T\Prob\vec{v}$.
Following \citet{singh2018fairness}, the fair ranking optimization can be expressed as a linear programming problem: 
\begin{align}
    &\max_{\Prob \in \Delta \cap \Gamma_\text{fair}}  \; \uvec^T\Prob\vvec 
    \quad \mbox{ where: } \quad \Delta : \Prob\vec{1}=\Prob^\top \vec{1} = \vec{1}, \;\; 
     \Prob_{j,k} \geq 0, \;\; 
     \forall_{1 \leq j,k \leq M}
\end{align}
and $\Gamma_\text{fair}$ denote any linear constraint set of the form $\fvec^\top\Prob\gvec = h$. Choosing $\fvec$ as the utility of items according to groups and $\gvec$ as the exposure of ranking position, enforces equality of exposure across protected groups.  
In contrast to \citet{singh2018fairness}, which uses this framework to re-rank the items to satisfy fairness constraints (i.e., a post-processing method),  we extend this linear perspective to derive a \emph{learning-to-rank} approach that learns to optimize utility and fairness simultaneously during training (i.e., an in-processing method).   
\subsection{Learning to Rank using Adversarial Approach}

We adopt a distributionally robust approach to the LTR problem, by constructing an worst-case adversary distribution on item utilities. 
We formulate the robust fair ranking construction as a minimax game between two players: 
a fair predictor $\Prob$ that makes a probabilistic  prediction over the set of all possible rankings to maximize expected ranking utility; and an adversary $\qq$ that approximates a probability distribution for the utility of items which minimizes the expected ranking utility. The adversary is additionally constrained to match the
feature moments of the empirical training distribution. 
Since we solve the problem for a given query, the query-dependent terms are omitted from the formulation for simplicity.
Additionally, we consider binary relevance values for items which results in binary utility in our formulation. 

In what follows we represent ranking items $d$ by their feature representation $\Xvec \in \Rbb^{M \times L}$ as a matrix of $M$ items with $L$ features. For a given item set $\Xvec$, the expected ranking utility of probabilistic ranking $\Pvec$ against the worst-case utility distribution $\qvec$ can be expressed as:
\begin{align}
\mathrm{U}({\bf X},\Prob,\qq) &=
\sum_{j=1}^{M}\Ebb_{u_j|\Xvec \sim \qvec} \left[ u_j \Ebb_{\pi_j|\Xvec \sim \Prob } \left[\expsr_{\pi_j}\right] \right] \label{eq:util} \\
    &= \sum_{j=1}^{M} \sum_{k=1}^{M} \qq(u_j=1|{\bf X})\Prob(\pi_j=k|{\bf X})\expsr_k =  \qq^\top \Prob \bv. \label{eq:util_binary}
\end{align}
 
Note that the transition from \eqref{eq:util} to \eqref{eq:util_binary} is possible in case of binary relevance which is the focus of our experiments. However expanding the expectation over $u_j$ in \eqref{eq:util} for multiple values is straightforward.  

The utility-maximizing optimization problem under fairness constraints can be formulated as follows:

\begin{definition}\label{def:fair_adversarial_ranking}
Given a training dataset of $N$ ranking problems $\Dcal = \{(\Xvec^i, \uvec^i)\}_{i=1}^N$, with $\uvec \in \Rbb^M$ being the true relevance and $\Xvec \in \Rbb^{M \times L}$ the feature representation of ranking problem of size $M$. The fair probabilistic ranking $\Pvec(\pi) \in \Rbb^{M \times M}$ in adversarial learning-to-rank learns a fair ranking that maximizes the worst-case ranking utility approximated by an adversary $\qvec(\vec{\uchk})$, constrained to match the feature statistics of the training data.
\begin{align}
& \max_{\Prob(\pi|\Xvec) \in \Delta \cap \Gamma_\text{fair}}\;\min_{\qvec(\vec{\uchk}|\Xvec)}
  \; 
  \Ebb_{\Xvec \sim \Ptilde}
  \left[ \mathrm{U} (\Xvec, \Pvec, \qq)\right]
    \label{eq:obj} \\
     \mbox{ s.t. } & \; \Ebb_{\Xvec \sim \tilde{P}} \left[\sum_{j=1}^M \Ebb_{\uchk_j|\Xvec \sim \qq} \left[\uchk_j \Xvec_{j,:} \right]\right] = \Ebb_{\Xvec, \uvec \sim \Ptilde}\left[
     \sum_{j=1}^M u_j \Xvec_{j,:}
     \right]
     \label{eq:const1} 
\end{align}

where $\Ptilde$ denotes the empirical distribution over ranking dataset $\Dcal = \{(\Xvec^i, \uvec^i)\}_{i=1}^N$, $\vec{\uchk}$ denotes the random variable for adversary relevance, and $\Delta$ denotes the set of doubly stochastic matrices. 
\end{definition}

This adversarial formulation has been utilized to provide fair and robust predictions under covariate shift \cite{rezaei2021robust} as well as for constructing reliable predictors for fair log loss classification \cite{rezaei2020fairness}. Similar to this line of work, our proposed approach imposes fairness constraints on predictor $\Prob$.

\section{Fairness of Exposure in Ranking}
There have been a variety of metrics introduced to quantify fairness in ranking. In probabilistic ranking, various fairness notions can be defined in the form $\vec{f}^\top\Prob\vec{g} = h$. The vector $\vec{f}$ is typically used to encode group membership and/or relevance of each document, whereas vector $\vec{g}$ represents the importance of a given position (e.g. position bias) and $h$ is scalar.  By choosing appropriate $\vec{f}$, $\vec{g}$ and $h$ wide range of fairness constraints like demographic parity, disparate treatment,
disparate impact can each be applied (separately) to the optimization framework \cite{singh2018fairness}.
Although our approach is flexible to implement all these fairness constraints, we focus on demographic parity in our exposition and experiments. 

Demographic parity requires equal exposure among different groups, $G_s$. Groups are defined based on the sensitive attribute (e.g., gender, race) of the items. 
The exposure of an item is measured based on its position in the ranking. A position bias function like $\expsr_k = \frac{1}{log(1+k)}$ determines the exposure of an item $j$ ranked in position $k$, i.e., $\pi_j = k$. Under a probabilistic ranking $\Prob$ the exposure for item $d_j$ is defined as  $\exposure{d_j} = \sum_{k=1}^M \Prob_{j,k} \expsr_k \label{exposure}$. The exposure for a group is the average exposure of items in that group, i.e., $\exposure{G_s}=\frac{1}{|G_s|}\sum_{d_j \in G_s} \exposure{d_j}$. Following \citet{singh2018fairness} to ensure demographic parity, members are shown for a query so that various groups receive a fair amount of exposure. Mathematically, this can be written as:
\begin{align}
   \frac{1}{|G_s|}\sum_{d_j \in G_s} \sum_{k=1}^M \Prob_{j,k} \expsr_k = \frac{1}{|G_{s'}|}\sum_{d_j \in G_{s'}} \sum_{k=1}^M \Prob_{j,k} \expsr_k = \tau, \:\:\ \forall s,s' \in S
   \label{eq:dp1}
\end{align}
where $S$ is the set of protected attributes. Based on (\ref{eq:dp1}), it follows that, in a fair ranking, average exposure of each group must be equal to average exposure of union of all groups in the ranking, i. e:
\begin{align}
   \frac{1}{|G_s|}\sum_{d_j \in G_s} \sum_{k=1}^M \Prob_{j,k} \expsr_k = \frac{1}{M} \sum_{j=1}^M \sum_{k=1}^M \Prob_{j,k} \expsr_k = \frac{1}{M} \sum_{k=1}^M \expsr_k \sum_{j=1}^M  \Prob_{j,k} = \frac{1}{M} \sum_{k=1}^M \expsr_k, \:\:\ \forall s \in S
   \label{eq:dp2}
\end{align}
Given that for all queries position bias function and number of items are fixed, $\frac{1}{M} \sum_{k=1}^M \expsr_k = \tau$ is a scalar. For each group $s$ we can write the fairness constraint in a vectorized form as:  
\begin{align}
  \frac{1}{|G_s|}\sum_{d_j \in G_s} \sum_{k=1}^M \Prob_{j,k} \expsr_k = \tau
    \Leftrightarrow & \:\:\sum_{d_j \in \D} \sum_{k=1}^M \Big(\frac{a_j}{|G_s|} \Big) \Prob_{j,k} \expsr_k = \tau, 
    & \:\:\sum_{d_j \in \D}a_j = |G_s| \\
    \Leftrightarrow & \:\:\mathbf{f}_s^\top P\bv = \tau \tag{with $\mathbf{f}_{s,j} = \frac{a_j}{|G_s|}$}
    \label{eq:dp3}
\end{align}
where $a_j$ is zero for items not in group $G_s$. Instead of choosing equal values ($a_j=1$) for items in group $G_s$,  we assign higher values for items with higher corresponding predicted utility. 
When we optimize for maximum fairness, this technique makes our approach retain higher utility and provide a preferable fairness-utility trade-off. 
In contrast, \citet{singh2018fairness} assigns equal values for every item in a group in the f vector (i.e., $\frac{1}{|G_s|}$ or $\frac{-1}{|G_s|}$ based on the group membership). 
This setup makes vector $f$  ignore the utility of items within the group.

By employing this technique, our approach is capable of providing a ranking sample based on Hungarian algorithm or an expected ranking sample based on Birkhoff-von Neumann method \cite{birkhoff1940lattice}.

\section{Optimization}

We solve the constrained minimax formulation in Definition \ref{def:fair_adversarial_ranking} in Lagrangian dual form, where we optimize the dual parameters $\thetavec \in \Rbb^{L \times 1}$ for the feature matching constraint of $L$ features by gradient decent. Rewriting the optimization in matrix notation yields: 
\begin{align}
\max_{\theta}& \; \Ebb_{\xvec,\uvec \sim \Ptilde}\Big[\max_{\Prob \in \Delta} \min_{0 \leq \qq \leq 1} \; \qq^\top \Prob \bv + \langle \qq - \uvec, {\sum}_l \theta_l \Xvec_{:,l} \rangle \Big] 
\mbox{ s.t. } \quad 
          \fvec_s^\top\Prob\bv\!=\!\merit, \quad s \in S,
          \label{eq1}
\end{align}

where $\Prob(\pi) \in \Rbb^{M \times M}$ is a doubly stochastic matrix, and the value of cell $\Pvec_{j,k}$ represents the probability that $\pi_j = k$. $\uvec \in \Rbb^{M \times 1}$ is a vector of true labels whose $j^\text{th}$ values is $1$ when the item $j$ is relevant to the query, i. e. $u_j = 1$ and 0 otherwise. $\qq \in \Rbb^{M \times 1}$ is a probability vector of the adversary's estimation of each item being relevant, and $\Xvec_{:,l} \in \Rbb^{M \times 1}$ denotes the $l$-th feature of $M$ samples. $S$ is the set of protected attributes and $\vvec \in \Rbb^{M \times 1}$ is a vector containing the values of position bias function for each position. To denote the Frobenius inner product between two matrices $\langle.,.\rangle$ is used, i.e., $\langle A,B\rangle = \sum_{i,j}A_{i,j}B_{i,j}$.

For optimization purposes, using strong duality, we push the maximization over $\qq$ to the outer-most level in \eqref{eq1}. Since the objective is non-smooth, for both $\Prob$ and $\qq$, we add strongly convex prox-functions to make the objective smooth. Furthermore, to make our approach handle feature sampling error, we add a regularization penalty to the parameter $\thetavec$. To apply \eqref{eq1} on training data, we replace empirical expectation with an average over all training samples. The new formulation is as follows:

\begin{align}
&\min_{\{ 0 \leq \qq^i \leq 1 \}_{i=1}^N} \max_{\theta} \frac{1}{N}\sum_{i=1}^{N}   \max_{\Prob^i \in \Delta}  \Big[ \qq^{i^\top} \Prob^i \bv^i - \langle \qq^i - \uvec^i, \tiny{\sum}_l \theta_l \Xvec^i_{:,l} \rangle \label{opt1} \\ & + \lambda \sum_{s \in S}\fvec^{i^\top}_{s} \Prob^i\vvec^i  -\frac{\mu}{2}\left \| \Prob^i \right \|^2_F    + \frac{\mu}{2}\left \| \qq^i \right \|^2_2 \Big] - \frac{\gamma}{2}\left \| \vec{\theta} \right \|^2_2 \notag 
\end{align}

where superscript $i$ corresponds to the $i^{th}$ sample from $N$ ranking problems in the training set. We denote $\lambda$, $\gamma$ and $\mu$ as the fairness penalty parameter, a regularization penalty parameter and a smoothing penalty parameter, respectively.  

The inner minimization over $\Prob$ and $\thetavec$ can be solved separately, given a fixed $\qq$. The minimization over $\thetavec$ has a closed-form solution where the $l^{th}$ element of $\thetavec^*$ is:

\begin{align}
    \theta^*_l = -\frac{1}{\gamma N}\sum_{i=1}^{N}\langle \qq^i - \uvec^i,   \Xvec^i_{:,l} \rangle.
    \label{theta}
\end{align}
Independently from $\thetavec$, we can solve the inner minimization over $\Prob$ for every training sample using a projection technique. The optimal $\Prob$ for $i^{th}$ training sample (i.e., $\Pvec^{i^*}$) is: 

\begin{align}
    \Prob^{i^*} =&  \argmax_{\Pvec^i \in \Delta 
    } \;
    \qq^{i^\top} \Prob^i \bv^i  + \lambda \fvec^{i^\top}\Prob^i\bv^i - \frac{\mu}{2}\left \| \Prob^i \right \|^2_F \notag \\
    \Prob^{i^*}=& \argmin_{\Pvec^i \in \Delta 
    } \;
    \frac{\mu}{2}\left \| \Prob^i - \frac{1}{\mu}(\qq^i + \lambda\fvec^i)\bv^{i^\top} \right \|^2_F - \frac{1}{2\mu}\left \| \qq^i\bv^{i^\top} \right \|^2_F.
    \label{pi_optimal}
\end{align}

As derived in \eqref{pi_optimal}, the minimization takes the form of $\min_{\Prob \geq 0} \left \|\Prob - \vec{R} \right \|^2_F,  \mbox{s.t.} :  \Prob \Vec{1} = \Prob^\top \Vec{1} = \vec{1}$, that is, we can interpret this minimization as projecting matrix $\frac{1}{\mu}(\qq^i + \lambda \fvec)\bv^{i^\top}$ into the set of doubly-stochastic matrices. The projection from an arbitrary matrix $\vec{R}$ to the set of doubly-stochastic matrices can be solved using ADMM projection algorithm. For proofs and further details, we refer the interested reader to the appendix.

This series of updates is run until the stopping conditions are met. For both the projected Quasi-Newton algorithm for optimizing $\qvec$ \eqref{opt1} and in the inner optimization for minimizing $\Prob$ \eqref{pi_optimal}, we utilize the same ADMM projection algorithm. Following  \citet{boyd2011distributed},  we apply stopping conditions based on the primal and dual residual optimality conditions.

\subsection{Inference} 
For prediction, we use $\thetavec$ and $\mu$ learned from training data while performing optimization in \eqref{opt1}. By removing constant terms in optimization we solve a similar optimization problem for test data. That is: 
{\small
\begin{align}
\min_{\{0 \leq \qq^i \leq 1\}_{i=1}^{N_\text{test}}} \frac{1}{N^\text{test}} \, \sum_{i=1}^{N^\text{test}}  \, \max_{\Prob^i \in \Delta} \, \Big[ \qq^{i^\top} \Prob^i \bv^i - \langle \qq^i , \tiny{\sum}_l \theta^*_l \Xvec^i_{:,l} \rangle + \lambda \sum_{s \in S}\fvec^{i^\top}_s \Prob^i \bv^i  -\frac{\mu}{2}\left \| \Prob^i \right \|^2_F + \frac{\mu}{2}\left \| \qq^i \right \|^2_2 \Bigg] 
\label{opt_inference}
\end{align}}%
where superscript $i$ pertains to the $i^{th}$ ranking problem in the the test set of size $N^\text{test}$. We follow the steps for solving the optimization in training. 
Although we play a minimax game between predictor and adversary in inference, we emphasize that there is no gradient learning for training data ($\theta$), and true relevance labels ($\uvec$) are not used in inference.
After convergence, we use the resulting $\Prob^*$ from the optimization to predict the ranking of items in the test set. We employ the Hungarian algorithm \cite{kuhn1955hungarian} to solve the problem of matching items to positions, incurring $\mathcal{O}(n^3)$ time. This results in a deterministic ranking function unlike other LTR methods with probabilistic inference. 

\section{Experiments}
\label{sec:exps}

In this section, we apply our fair adversarial ranking framework to the task of \emph{learning to rank} with \emph{group fairness} constraints. The learning task is to determine the feature function in the training based on the items' ground truth utilities and fairness constraints. At testing time, 
this feature function coupled with a penalty for fairness violation is used to determine the ranking for the items in the test set with maximum utility while satisfying fairness constraints. 
\subsection{Setup}
\begin{wraptable}{r}{0.45\textwidth}
\caption{Dataset characteristics.}
\label{tab:datasets}
\vspace{2mm}
\centering
\begin{tabular}{lrrl}
\hline
Dataset    & \multicolumn{1}{c}{$n$} & \multicolumn{1}{c}{Features} & \multicolumn{1}{c}{Attribute} \\ \hline
{\tt Adult}      & 45,222                  & 12       & Gender    \\
{\tt COMPAS}     & 6,167                   & 10       & Race      \\
{\tt German}     & 1,000                   & 20       & Gender    \\
\hline
\end{tabular}
\end{wraptable}
To evaluate the effectiveness of our approach, we perform experiments on three benchmark datasets. Following \citet{singh2019policy}, we construct a learning-to-rank task from a classification dataset. First, we split dataset randomly to a disjoint train and test set. Then from each train/test set we construct a corresponding learning-to-rank train/test set. For each query, we sample randomly with replacement a set of 10 candidates each, representative of both relevant and irrelevant items, where, on average four individuals are relevant. Each individual in the candidate set is a member of a group $G_s$ based on it's protected attribute. The training data consists of 500 ranking problems. We evaluate our learned model on 100 separate ranking problems serving as the test set. We repeat this process 10 times and report the 95\% confidence interval in the results. The regularization constant $\gamma$ and smoothing penalty parameter $\mu$ in (\ref{opt1}) are chosen by 3-fold cross validation. We describe datasets used in our experiments:

\begin{itemize}
    \item UCI {\tt Adult}, census income dataset \cite{Uci2017}.
     The goal is to predict whether income is above \$50K/yr on the basis of census results.
    \item The {\tt COMPAS} criminal recidivism risk assessment dataset \cite{larson2016we} is designed to predict whether a defendant is likely to reoffend based on criminal history.
    \item UCI {\tt German} dataset \cite{Uci2017}. Based on personal information and credit history, the goal is to classify good and bad credit.
\end{itemize}
Table \ref{tab:datasets} shows the statistics of each dataset with their protected attributes.

\paragraph{Baseline methods}
\label{baselines}
To evaluate the performance of our model, we compare it against three different baselines that have similarities to and differences from our model: FAIR-PGRank \cite{singh2019policy} and DELTR \cite{zehlike2020reducing} are in-processing, LTR methods, like ours; the Post-Processing method of \citet{singh2018fairness} employs the fairness constraint formulation that we build our optimization framework based on. 
We also add a Random baseline that ranks items in each query randomly to give context to NDCG. We discuss baseline methods in more details\footnote{For all baseline methods we use the implementation from {\tt \url{https://github.com/ashudeep/Fair-PGRank}}.}:

\begin{itemize}
    
    \item {\bf Post Processing} ({\sc Post\_Proc}) \cite{singh2018fairness}
    In order to make a fair comparison with in-processing LTR approaches, we first learn a linear regression model that is trained on all query-item sets in the training data and predicts the relevance of an item to a query in test set. Then, these estimated relevances are used as input to the linear program optimization described in \citet{singh2018fairness} with a demographic parity constraint for group fairness.  
    \item {\bf Fair Policy Ranking} ({\sc Fair\_PGRank}) \cite{singh2019policy} An end-to-end, in-processing LTR approach that uses a policy gradient method, directly optimizing for both utility and fairness measures.
    \item {\bf Reducing Disparate Exposure} ({\sc DELTR}) \cite{zehlike2020reducing} An in-processing LTR method that optimizes a weighted summation of a loss function and a fairness criterion. The loss function is a Cross Entropy loss designed for ranking \cite{cao2007learning} and fairness objective is a squared hinge loss based on disparate exposure.  
    
\end{itemize}

\textbf{Evaluation Metrics} We use the \emph{normalized discounted cumulative gain} (NDCG) \cite{jarvelin2002cumulated}, as the utility measure. This is defined as: 
\begin{align}
\label{eq:ndcg}
\text{\emph{NDCG}} = \frac{1}{Z} \sum_{j=1}^M \frac{2^{\rel(d_j)}-1}{\log(1+\pi_j)},
\end{align}
where 
$Z$ is the DCG for ideal ranking and is used to normalize the ranking so that a perfect ranking would give a NDCG score of $1$.

For the fairness evaluation in our approach we use \emph{demographic parity} as our fairness violation metric which is based on disparity of average exposure across two groups:
\begin{equation}
    \hat{D}_{group}(\Prob) = |\exposure{G_0} - \exposure{G_1}|.
    \label{eq:31}
\end{equation}

\begin{figure*}[t]

    \includegraphics[width=.33\textwidth]{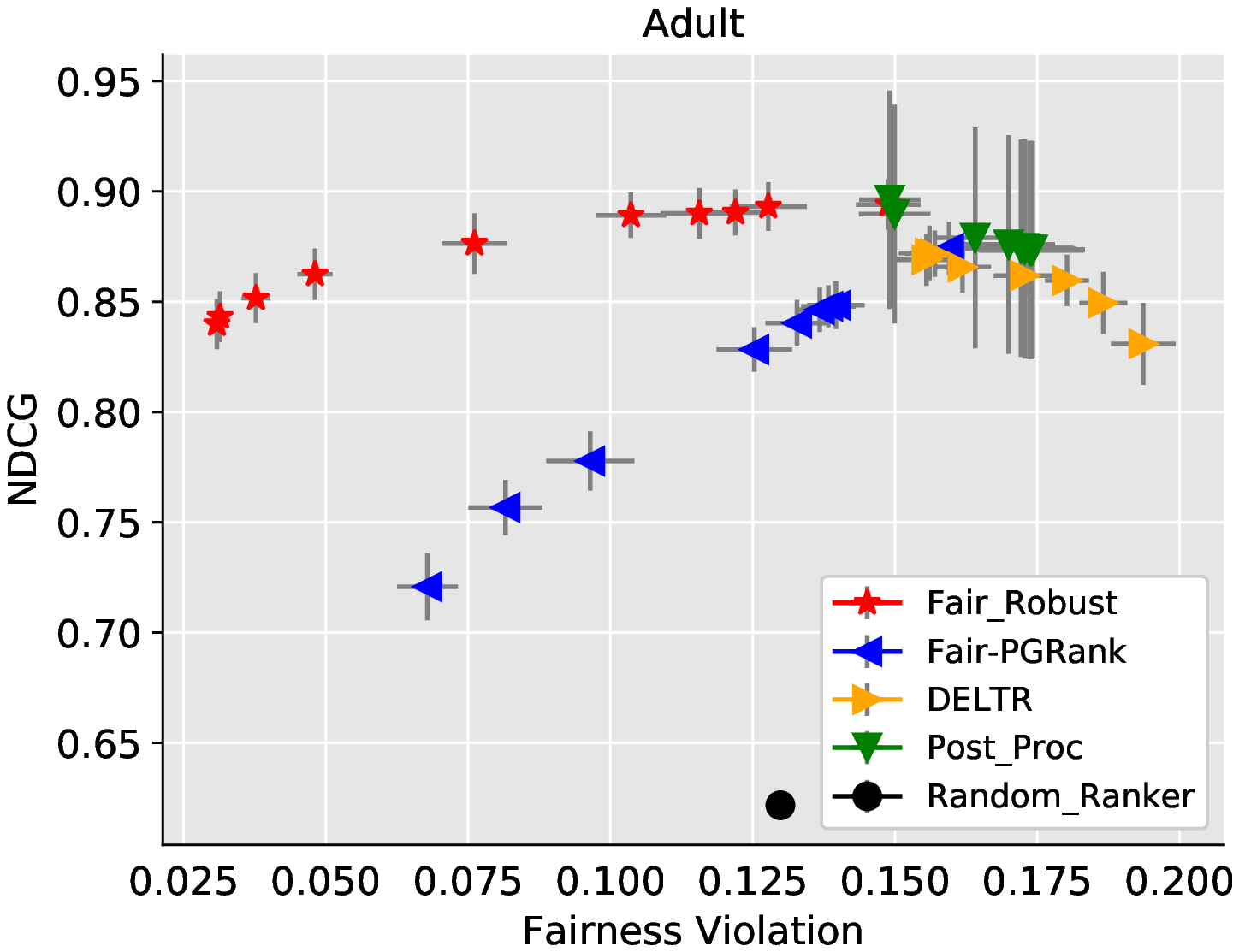} 
    \includegraphics[width=.33\textwidth]{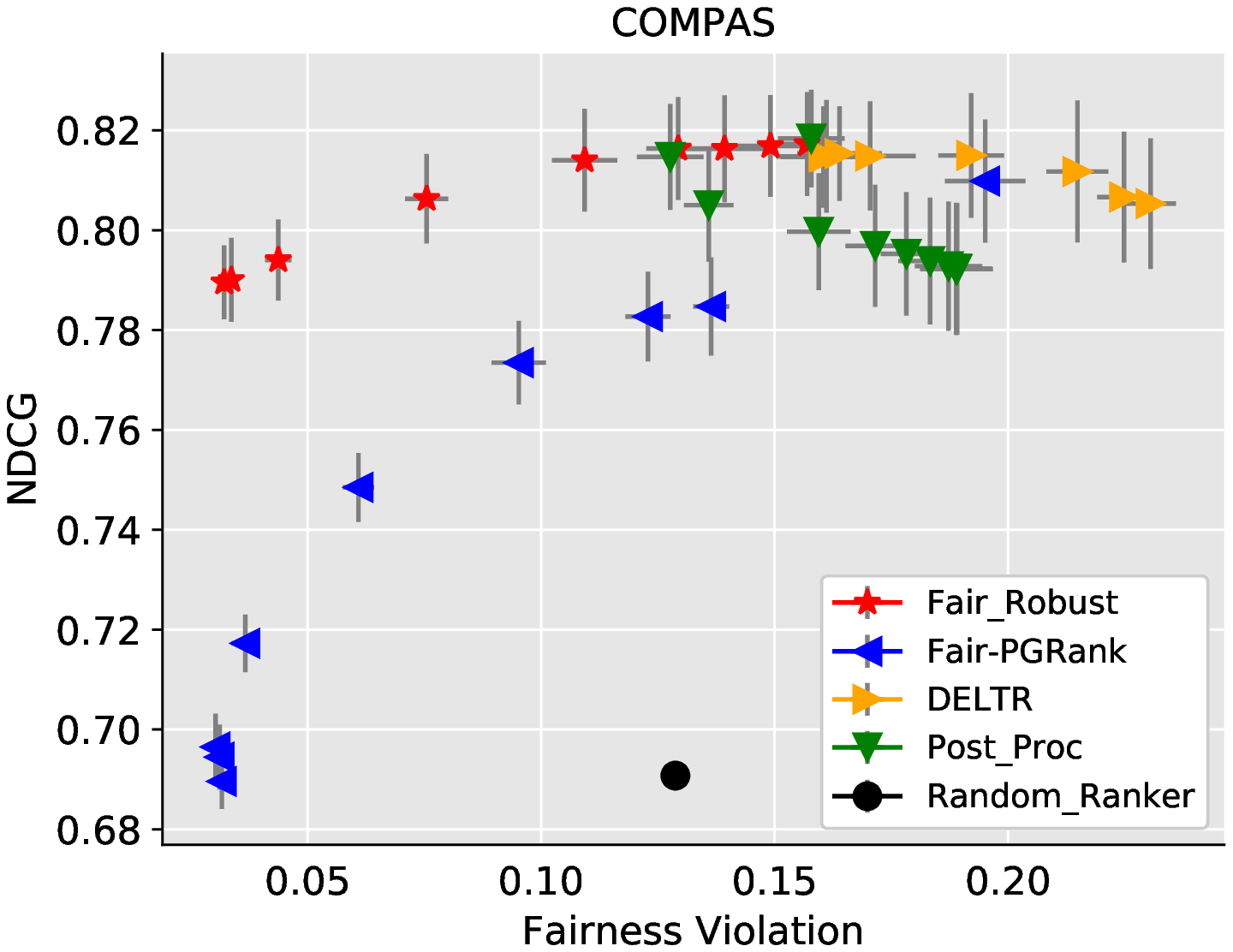}
    \includegraphics[width=.33\textwidth]{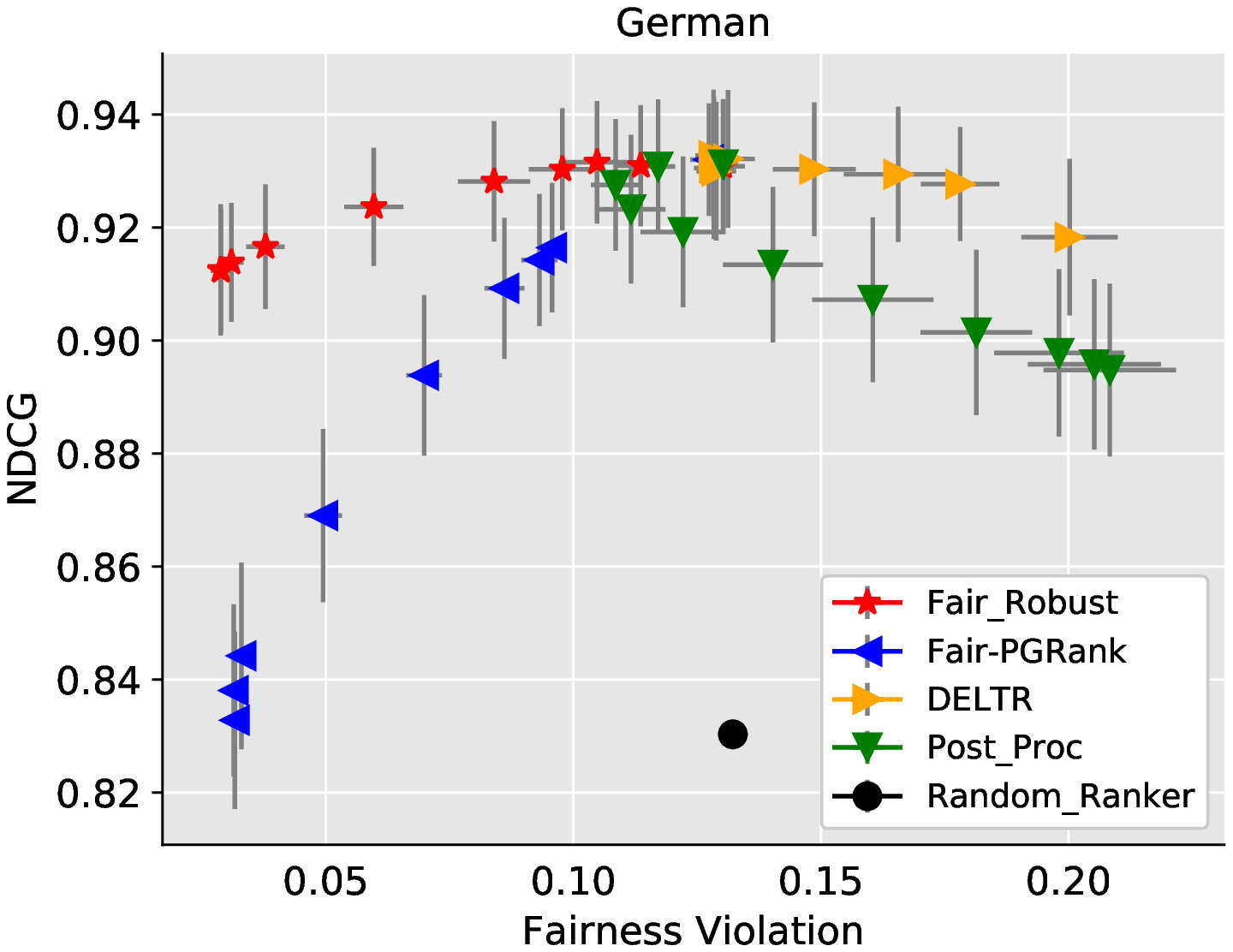}
 
    \caption{Average \emph{NDCG} versus average \emph{difference of demographic parity} (DP) on test samples, for increasing degrees of fairness penalty $\lambda$ in each method. Fair\_Robust: $\lambda \in [0, 10]$, Fair\_PGRank: $\lambda \in [0, 10]$, DELTR: $\lambda \in [0, 10^6]$, Post\_Proc: $\lambda \in [0, 0.2]$.}
    \label{fig:results}

\end{figure*}
\subsection{Results}
\label{results}
Figure \ref{fig:results} shows the performance of our model against baselines on the three benchmark datasets. We observe a trade-off between fairness and utility in both  {\sc Fair-PGRank} and {\sc Fair-Robust} (our approach), i.e., as we increase the fairness penalty parameter ($\lambda$), demographic parity difference (as a measure of fairness violation) and NDCG both drop.
While {\sc DELTR} and {\sc Post-Proc} achieve comparable NDCG when $\lambda = 0$, they fail to satisfy demographic parity as we increase $\lambda$ and are unable to provide a sufficient utility-fairness trade-off when high levels of fairness are desired. 

In all three datasets {\sc Fair-Robust} outperforms {\sc Fair-PGRank} in terms of ranking utility when fairness is a priority. When comparing the utility-fairness trade-off between two approaches we observe that {\sc Fair-Robust} can retain higher NDCG in high levels of fairness and provides a preferable trade-off.

To evaluate how fair these methods can perform, we find the minimum fairness violation that occurs when we run baseline methods on each dataset. We report the minimum $\hat{D}_{group}$ and corresponding NDCG resulting for different values of $\lambda$. Table \ref{tab:fairest} shows the results for this evaluation. We highlight the fairest results achieved by these methods. Since one main goal in this paper is to retain high NDCG when minimum fairness violation achieved we highlight highest NDCG values that their corresponding fairness violation is below a threshold ($\hat{D}_{group} < 0.1$). 
{\sc Fair-Robust} outperforms all baseline methods in terms of fairest results and NDCG.
\begin{table*}[th]
\vspace{-1mm}
  \caption{
        Fairest results achieved by {\sc Fair-Robust} and baseline methods on benchmark datasets.
    }
    \label{tab:fairest}
    \vspace{1mm}
    \centering
    \small
    \resizebox{\textwidth}{!}{
    \begin{tabular}{cccccccc}
        \toprule
        &
        \multirow{2}{*}{\diagbox{Method}{Dataset}}
        &
        \multicolumn{2}{c}{\parbox[c]{3.6cm}{\centering \small {\tt Adult}}} 
        & 
        \multicolumn{2}{c}{\parbox[c]{3.6cm}{\centering \small {\tt COMPAS}}} 
        & 
        \multicolumn{2}{c}{\parbox[c]{3.6cm}{\centering \small {\tt German}}} 
        \\
        \cmidrule(lr){3-4} \cmidrule(lr){5-6} \cmidrule(lr){7-8} 
        
        &    & 
        NDCG & $\hat{D}_{group}$ &
        NDCG & $\hat{D}_{group}$ & 
        NDCG & $\hat{D}_{group}$  \\
        \midrule
        \parbox[t]{2mm}
        & {\sc Fair\_Robust} & \bf{0.839 $\pm$ 0.011} & \bf{0.031 $\pm$ 0.001} & \bf{0.789 $\pm$ 0.007} & 0.032 $\pm$ 0.001 & \bf{0.912 $\pm$ 0.011} & \bf{0.029 $\pm$ 0.002}  \\
        & {\sc Fair\_PGRank} & 0.721 $\pm$ 0.015 & 0.068 $\pm$ 0.005 & 0.696 $\pm$ 0.006 & \bf{0.030 $\pm$ 0.001} & 0.838 $\pm$ 0.015 & 0.031 $\pm$ 0.001 \\ 
        & {\sc DELTR} & 0.869 $\pm$ 0.011 & 0.155 $\pm$ 0.005 & 0.815 $\pm$ 0.010 & 0.160 $\pm$ 0.009 & 0.933 $\pm$ 0.011 &	0.128 $\pm$ 0.003  \\
        & {\sc Post\_Proc} & 0.896 $\pm$ 0.049 & 0.149 $\pm$ 0.005 & 0.818 $\pm$ 0.009 & 0.158 $\pm$ 0.007 & 0.927 $\pm$ 0.011 & 0.108 $\pm$ 0.005  \\
        \bottomrule
    \end{tabular}
    }
\end{table*}
\vspace{-3mm}
\section{Conclusions}

\label{conclusion}
In this paper, we developed a new LTR system that achieves fairness of exposure for protected groups while maximizing utility to the users. 

Our adversarial approach constructs a minimax game with the ranker player choosing a distribution over rankings constrained to provide fairness while maximizing utility and an adversary player choosing a distribution of item relevancies that minimizes utility while being similar to training data properties.  
We show that our method is able to trade-off between utility and fairness much better at high levels of fairness than existing baseline methods. 
Our work addresses the problem of providing more robust fairness given a chosen fairness criterion, but does not answer the broader question of which fairness criterion is appropriate for a particular ranking application. 
Since optimizing one fairness criterion can be detrimental to other fairness criteria, this is an important practical consideration with societal implications.
More extensive evaluation based on incorporating other fairness metrics, such as disparate treatment and using real-world ranking datasets are both important future directions.

\section*{Acknowledgements}
This work was supported by the National Science Foundation Program on Fairness in AI in collaboration with Amazon under award No. 1939743.

\section*{Broader Impact}

The social implications of rankings go beyond their immediate utility, since higher rankings provide opportunities for individuals and groups associated with ranked items. As a consequence, biases in ranking systems---whether intentional or not---raise ethical concerns about their long-term economic and societal harming effect.
This work offers an approach for robustly seeking fair rankings and could be of general benefit to individuals impacted by alternative systems that suffer from existing biases. This work addresses the problem of providing more robust fairness given a chosen fairness criterion, but does not answer the broader question of which fairness criterion is appropriate for a particular ranking application. 

\bibliography{references}
\bibliographystyle{unsrtnat}

\appendix
\input{appendix.tex}

\end{document}

%% file: appendix.tex
\newpage
\appendix

\section{Appendix}

\subsection{Doubly-stochastic Matrix Projection}

The projection from an arbitrary matrix $\vec{R}$ to the set of doubly-stochastic matrices can be formulated as: 
\begin{align}
    \min_{\Prob \geq 0} \left \|\Prob - \vec{R} \right \|^2_F, 
    \quad \mbox{s.t.} : & \Prob \Vec{1} = \Prob^\top \Vec{1} = \vec{1}
\end{align}

We divide the doubly-stochastic matrix constraint into two sets of constraints $C_1 : \Prob \Vec{1}= \vec{1}$ and $\Prob \geq \vec{0}$, and $C_2 : \Prob^\top \Vec{1}= \vec{1}$ and $\Prob \geq 0$. Using this construction, we convert the optimization above into ADMM as follows:

\begin{align}
    \min_{\Prob,\vec{S}}  \frac{1}{2}&\left \|\Prob - \vec{R} \right \|^2_F + \frac{1}{2}\left \|\vec{S} - \vec{R} \right \|^2_F + \vec{I}_{C_1}(\Prob) + \vec{I}_{C_2}(\vec{S})\notag\\
    \quad  \mbox{s.t.} & :  \Prob - \vec{S} = \vec{0}
\end{align}

The augmented Lagrangian for this optimization is:

\begin{align}
    \mathcal{L}_\rho(\Prob,\vec{S},\vec{W}) = & \frac{1}{2}\left \|\Prob - \vec{R} \right \|^2_F + \frac{1}{2}\left \|\vec{S} - \vec{R} \right \|^2_F + \vec{I}_{C_1}(\Prob)\notag\\& + \vec{I}_{C_2}(\vec{S}) + \frac{\rho}{2}\left \|\Prob - \vec{S} + \vec{W} \right \|^2_F,
\end{align}

Where $\rho$ is the ADMM penalty parameter and $\vec{W}$ is the scaled dual variable. From the augmented Lagrangian, we compute the update for $\Prob$ as:

\begin{align}
    \Prob^{t+1} & =  \argmin_{\Prob}\mathcal{L}_\rho(\Prob,\vec{S}^t,\vec{W}^t) \notag\\
    & =\argmin_{\{\Prob \geq 0|\Prob \Vec{1} = \vec{1}\}} \frac{1}{2}\left \|\Prob - \vec{R} \right \|^2_F + \frac{\rho}{2}\left \|\Prob - \vec{S}^t + \vec{W}^t \right \|^2_F \notag\\
    & =\argmin_{\{\Prob \geq 0|\Prob \Vec{1} = \vec{1}\}} \left \|\Prob - \frac{1}{1 + \rho} (\vec{R} + \rho(\vec{S}^t - \vec{W}^t) )\right \|^2_F.
\end{align}

The minimization above can be interpreted as a projection to the set $\{\Prob \geq 0|\Prob\vec{1}=\vec{1}\}$, which can be realized by projecting to the probability simplex independently for each row of the matrix $\frac{1}{1 + \rho} (\vec{R} + \rho(\vec{S}^t - \vec{W}^t))$. Similarly, the ADMM update for $\vec{S}$ can also be formulated as a column-wise probability simplex projection. The technique for projecting a point to the probability simplex has been studied previously, e.g., by \citet{duchi2008efficient}. Therefor, our ADMM algorithm consists of the following updates:

\begin{align}
    \Prob^{t+1} &=  \textrm{Proj}_{C_1}(\frac{1}{1 + \rho} (\vec{R} + \rho(\vec{S}^t - \vec{W}^t) ))\notag\\
    \vec{S}^{t+1} &= \textrm{Proj}_{C_2}(\frac{1}{1 + \rho} (\vec{R} + \rho(\vec{S}^t + \vec{W}^t) ))\notag\\ 
    \vec{W}^{t+1} &= \vec{W}^t + \Prob^{t+1} - \vec{S}^{t+1}.
\end{align}

Now that we have a general formulation for the projection step, we replace $\vec{R}$ with $\frac{1}{\mu}(\qq_i + \alpha_i\fvec)\bv_i^\top$, which we acquired when optimizing $\Prob_i^*$ in Eq. \eqref{pi_optimal}:

\begin{align}
    \Prob^{t+1} = & \textrm{Proj}_{C_1}(\frac{1}{1 + \rho} (\frac{1}{\mu}(\qq + \lambda\fvec)\bv^\top + \rho(\vec{S}^t - \vec{W}^t) ))\notag\\
    \vec{S}^{t+1} = & \textrm{Proj}_{C_2}(\frac{1}{1 + \rho} (\frac{1}{\mu}(\qq + \lambda\fvec)\bv^\top + \rho(\vec{S}^t + \vec{W}^t) ))\notag\\ 
    \vec{W}^{t+1} = & \vec{W}^t + \Prob^{t+1} - \vec{S}^{t+1}.
\end{align}

\subsection{Optimal $\Prob$ Derivation}

We can solve the inner minimization over $\Prob$ for every training sample using a projection technique. The optimal $\Prob$ for $i^{th}$ training sample (i.e., $\vec{P_i^*}$) is:
\begin{align}
    \Prob_i^* =&  \argmax_{\vec{P_i} \geq 0|\Prob_i^\top \Vec{1} = \vec{1}^\top\Prob_i = \vec{1}} \qq_i^\top \Prob_i \bv_i  + \lambda \fvec_i^\top\Prob_i\bv_i -\frac{\mu}{2}\left \| \Prob_i \right \|^2_F \notag \\
    =&  \argmin_{\vec{P_i} \geq 0|\Prob_i^\top \Vec{1} = \vec{1}^\top\Prob_i = \vec{1}}  \frac{\mu}{2}\left \| \Prob_i \right \|^2_F  - \qq_i^\top \Prob_i \bv_i  - \lambda \fvec_i^\top\Prob_i\bv_i \notag\\
    =&  \argmin_{\vec{P_i} \geq 0|\Prob_i^\top \Vec{1} = \vec{1}^\top\Prob_i = \vec{1}} \frac{\mu}{2}\left \| \Prob_i \right \|^2_F  - \operatorname{tr}{(\Prob_i \bv_i \qq_i^\top )} - \lambda \operatorname{tr}{(\Prob_i\bv_i\fvec_i^\top)} \notag\\
    =& \argmin_{\vec{P_i} \geq 0|\Prob_i^\top \Vec{1} = \vec{1}^\top\Prob_i = \vec{1}}
    \frac{\mu}{2}\left \| \Prob_i - \frac{1}{\mu}(\qq_i + \lambda\fvec_i)\bv_i^\top \right \|^2_F - \frac{1}{2\mu}\left \| \qq_i\bv_i^\top \right \|^2_F.
    \label{app:pi}
\end{align}